\documentclass{optica-article}

\journal{opticajournal} 

\articletype{Research Article}

\usepackage{amsmath,upgreek}

\DeclareGraphicsExtensions{.eps,.png}

\usepackage{titlesec}
\titlespacing\section{0pt}{1.5ex}{0.6ex}
\titlespacing\subsection{0pt}{1.5ex}{0.6ex}
\titlespacing\subsubsection{0pt}{1.5ex}{0.6ex}

\usepackage{algorithm}
\usepackage{algpseudocode}

\usepackage[labelformat=simple]{subcaption}

\DeclareCaptionLabelFormat{subcaptionlabel}{\normalfont(\textbf{#2}\normalfont)}
\usepackage{cleveref}
\crefname{figure}{Fig.}{Figs.}
\Crefname{figure}{Figure}{Figures}
\crefname{equation}{Eq.}{Eqs.}
\Crefname{equation}{Equation}{Equations}
\crefrangelabelformat{equation}{(#3#1#4)-(#5#2#6)}
\crefname{section}{Sec.}{Secs.}
\Crefname{section}{Section}{Sections}
\crefname{subsection}{Subsec.}{Subsecs.}
\Crefname{subsection}{Subsection}{Subsections}

\DeclareMathOperator*{\argmin}{\arg\!\min}

\begin{document}

\title{Noise2Ghost: Self-supervised deep convolutional reconstruction for ghost imaging}

\author{Mathieu~Manni,\authormark{1,2} Dmitry~Karpov,\authormark{3} Kees~Joost~Batenburg,\authormark{4} Sharon~Shwartz,\authormark{2} and Nicola~Viganò\authormark{3,*}}

\address{\authormark{1}Algorithms and scientific Data Analysis group, ESRF --- The European Synchrotron, Grenoble, 38000, France\\
\authormark{2}Physics Department and Institute of Nanotechnology and Advanced Materials, Bar Ilan University, Ramat Gan, 52900, Israel\\
\authormark{3}Université Grenoble Alpes, CEA, IRIG-MEM, Grenoble, 38000, France\\
\authormark{4}Leiden Institute of Advanced Computer Science, Leiden Universiteit, 2333, CA Leiden, The Netherlands}

\email{\authormark{*}nicola.vigano@cea.fr} 

\begin{abstract}
We present a new self-supervised deep-learning-based Ghost Imaging (GI) reconstruction method, which provides unparalleled reconstruction quality for noisy acquisitions among unsupervised methods. We present the supporting mathematical framework and results from theoretical and real data use cases.
Self-supervision removes the need for clean reference data while offering strong noise reduction. This provides the necessary tools for addressing signal-to-noise ratio concerns for GI acquisitions in emerging and cutting-edge low-light GI scenarios. Notable examples include micro- and nano-scale x-ray emission imaging, e.g., x-ray fluorescence imaging of dose-sensitive samples. Their applications include in-vivo and in-operando case studies for biological samples and batteries.
\end{abstract}



\section{Introduction}
\begin{figure}[t]
    \centering
    \begin{subfigure}[b]{0.49\linewidth}
        \includegraphics[width=\linewidth]{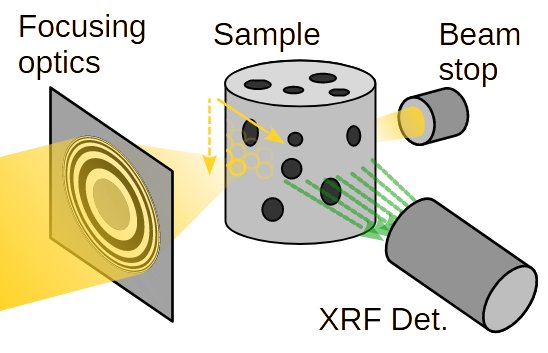}
        \subcaption{Pencil Beam}
        \label{fig:xrf-img:pb}
    \end{subfigure}
    \hfill
    \begin{subfigure}[b]{0.49\linewidth}
        \includegraphics[width=\linewidth]{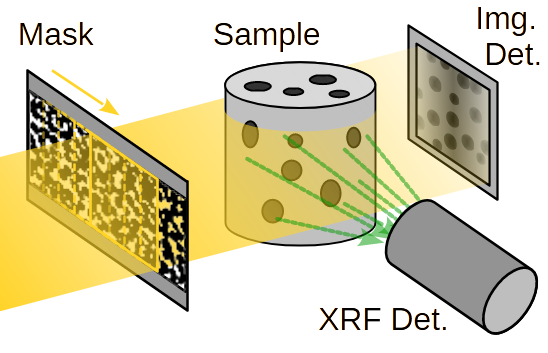}
        \subcaption{Classical Ghost Imaging}
        \label{fig:xrf-img:gi}
    \end{subfigure}

    \caption{Schematic representation of diffused emission signal acquisitions (e.g., x-ray fluorescence imaging) using pencil raster beam scanning~(\textbf{\subref{fig:xrf-img:pb}}) and ghost imaging~(\textbf{\subref{fig:xrf-img:gi}}). The former uses a point beam to scan every pixel to form an image, while the latter illuminates the sample with a series of structured beams.}
    \label{fig:xrf-imaging}
    \vspace{-0.2cm}
\end{figure}

Ghost imaging (GI) is a paradigm-changing technique that is mostly known and anticipated for its promise to cut radiation doses. However, these possible radiation dose advantages rely on a much more fundamental concept: The ability to choose the trade-off between spatial resolution, field-of-view (FoV), and acquisition time somewhat independently of the incident beam size. Traditionally, to spatially resolve diffused signals like x-ray fluorescence (XRF), one had to raster scan the object of interest with focused illumination (pencil-beam, PB, \cref{fig:xrf-img:pb}). The PB focal spot size determines the spatial resolution. Classical GI achieves that by probing extended regions of a sample at once with structured illumination (\cref{fig:xrf-img:gi}). This provides spatially resolved information, whose resolution is independent of the beam size and detector pixel size, but depends instead on the size of the beam structures. Hereafter, we will refer to classical GI simply as GI, which, unlike quantum GI~\cite{Pittman1995}, does not require pairs of quantum entangled photons.
\\
A detector that does not observe the illumination beam collects the signals of interest. The acquired GI realizations, composed of the detected signals with each associated illumination beam structure, are then combined and computationally reconstructed into a two-dimensional projection image of the probed object (in the given contrast).
This means that the acquired data mathematically resides in a different space from the reconstructed image space, and the reconstruction problem is a so-called \textit{inverse problem}.
The algorithm used to perform this inversion, known as the \textit{reconstruction method}, significantly influences the quality and accuracy of the reconstructed image, particularly under noisy conditions.
GI is independent of the contrast and probe type used, and it has been demonstrated with radiation across the whole electromagnetic spectrum~\cite{Katz2009,Pelliccia2016}, neutrons~\cite{Kingston2020}, electrons~\cite{Li2018}, and atoms~\cite{Khakimov2016}. Thus, it can be used for the same applications where PB raster scanning is used.
\\
The advantages of GI compared to PB acquisitions stem from the inherent small- and large-scale correlations found in natural images, which make them compressible under suitable representations~\cite{Brunton2019}, while non-structured noise is not.
Thanks to the structured illumination, which can capture unique large- and small-scale information for each measurement, GI grants image reconstruction with fewer data points than reconstructed pixels. Thus, GI unlocks potential gains in acquisition speed or deposited dose, compared to raster scanning, by simply reducing the number of acquired data points~\cite{Lane2020}.
\\
GI offers a specific advantage over PB scanning for dose-sensitive samples. High-flux focused PBs create significant excess charges in small localized regions at each exposure from radiation-induced local ionization of the samples. This is one of the main contributors to damage in biological samples and the skewing of functioning parameters in batteries during acquisitions. On the contrary, GI creates this same charge on a much larger area (i.e., the whole illumination field-of-view). The latter can recombine and/or disperse faster than the former, leading to lower degradation over time. For instance, in cryogenic electron microscopy (also known as cryo-EM), reduced dose rates have been observed to produce less damage compared to higher dose rates at the same total dose~\cite{Karuppasamy2011}.
\\
Thanks to all these advantages, GI has the potential to spark many breakthroughs in applications in various fields ranging from biomedical imaging to remote sensing.
However, GI has only seen limited applicability so far, especially in the above-mentioned nano-scale imaging applications. In these cases, the observed signals are affected by overwhelmingly high Poisson noise levels due to their very small photon fluxes. This noise type does not enjoy the intrinsic noise reduction of GI, known as Felgett's advantage, thus reducing GI's advantage over PB~\cite{Lane2020}.
Both traditional methods, based on regularized variational reconstructions (e.g., Total Variation minimization - TV-min~\cite{Rudin1992}), and modern machine learning (ML) techniques (both supervised~\cite{Rizvi2020} and unsupervised~\cite{Wang2022}) are not specifically designed for this scenario.
Here, we propose a GI reconstruction method particularly suited for working with acquisitions affected by high noise levels, regardless of whether the noise type is Gaussian or Poisson. Our method leverages the latest developments in unsupervised ML techniques, so it does not require high-quality reference data.
This work is a step towards practical applications of GI, particularly for photon-limited scenarios like nano-scale x-ray emission imaging~\cite{Li2023}.

\section{Method}
\label{sec:method}
In this section, we present the proposed method, Noise2Ghost (N2G). First, we introduce related methods. Then, we describe the signal formation (forward) model and present our proposed reconstruction method. Finally, we show how this method addresses random acquisition noise. The code of the algorithms presented in this section is available at \cite{Vigano2025}.

\subsection{Existing methods}
GI's ability to obtain images with more pixels than the acquired realizations results in an undersampled reconstruction problem. These reconstructions require some prior knowledge (regularization) to recover accurate images: Unregularized least-squares (LS) reconstructions present specific high-frequency large-scale artifacts that resemble high noise levels.
Methods employing convolutional neural networks (CNNs) are currently the best-performing tools for image processing applications. They can capture small- and large-scale correlations and learn the most suitable representations to capture the features of the treated signals (acting like regularizers)~\cite{Lempitsky2018}. Thus, they are also better suited for GI reconstructions than traditional variational methods, e.g., TV-min, which use strong but also rather simple \emph{a priori} assumptions on the types of features present in the reconstructed images.
\begin{table}[ht]
    \centering
    \caption{Comparison of GI Reconstruction Methods}
    \label{tab:comparison-methods}
    \raggedright
    \setlength{\tabcolsep}{3pt} 
    \small
    \begin{tabular}{|p{0.175\linewidth}||p{0.25\linewidth}|p{0.25\linewidth}|p{0.25\linewidth}|}
        \hline
        \textbf{Method} & \textbf{Description} & \textbf{Advantages} & \textbf{Limitations} \\
        \hline \hline
        Unregularized Least-Squares (LS) & Direct reconstruction without regularization & Simple, no prior assumptions & High-frequency artifacts; high noise levels; no compression \\
        \hline
        Regularized Variational (e.g., TV) & Uses strong \emph{a priori} assumptions (e.g., sparsity, smoothness) & Robust; cheap; reference in literature & Oversimplified assumptions; limited feature capture \\
        \hline
        Supervised denoising (e.g., DeepGhost) & Uses pairs of LS and high-quality reconstructions & Effective for known image types & Dependent on large, similar training data; sensitive to parameter variations \\
        \hline
        Self-Supervised denoising (e.g., N2V, N2N) & No pre-training; denoising in image space & No clean data required & Unable to deal with missing realization artifacts \\
        \hline
        Unsupervised (e.g., GIDC, INRs) & No pre-training; physics-informed learning & No clean data required; best results with high compression & Reduced quality with high noise; relies on regularization/early stopping \\
        \hline
    \end{tabular}
\end{table}
\\
Both ML methods and CNN architectures have seen an explosion in development and applications over the last decade. Quite a few methods have been specifically developed for GI reconstructions. Notable examples are supervised methods~\cite{Rizvi2020}, where a CNN (e.g., a U-net~\cite{Ronneberger2015}) compares LS GI reconstructions from undersampled acquisitions against their corresponding high-quality reconstructions. The network learns to identify and correct the corrupted features in the LS images.
This approach heavily depends on large amounts of known examples that present strong similarity to the images of interest. It also requires specific training data for each acquisition parameter variation (e.g., noise levels, image size, levels of undersampling, etc).
\\
Untrained generator-network methods do not require pre-training against high-quality reference images, thereby removing the need for large databases~\cite{Habring2024, Qayyum2022}. They can incorporate knowledge of the image formation model in the learning algorithm (e.g., physics-informed ML), and train the model to produce an image that fits the acquired data. A good example is the method called GIDC~\cite{Wang2022}, which is based on the deep image prior (DIP)~\cite{Lempitsky2018}. As a drawback, these methods exhibit lower reconstruction quality for high noise levels in the acquired data. They delegate noise reduction solely to regularization and early stopping.
Similar to DIP-based methods, implicit neural representations (INRs) -based methods use untrained multi-layer perceptron (MLP) NNs to reconstruct an image without high-quality reference data~\cite{Sitzmann2019, Li2023}. They accept pixel coordinates instead of input images. During training, the MLP learns a representation of the imaged object, which produces the desired image for the input coordinates.
\\
Self-supervised methods like Noise2Noise (N2N) use the same training procedure as supervised methods, but they use noisy data as target~\cite{Lehtinen2018}. They only require two noisy realizations of the same measurement (e.g., two noisy images of the same scene) and exploit the fact that the only difference between the two should be the noise. For each of the two measurements, they train the CNN to predict the other. As previously mentioned, CNNs can capture small and large-scale correlations, leading them to learn the features of the measurements while discarding their uncorrelated parts, i.e., the noise.
Similarly, single-image self-supervised denoising methods have also emerged, including Noise2Void~\cite{Krull2019}, and a few more recent developments~\cite{Lequyer2022, Huang2022, Li2025, Wu2025}. Compared to N2N, these methods only exploit the spatial correlation of imaging data.
N2N can be adapted to work with inverse problems like computed tomography~\cite{Hendriksen2020}, while single-image methods can be directly applied to least-squares reconstructions. However, as we will see in \cref{subsec:N2I}, they do not handle the so-called \textit{missing realization} problem, which refers to artifacts arising from not having enough realizations to represent the imaged objects, and that is usually an important source of noise in GI reconstructions.
A brief comparison of all the mentioned methods is given in \cref{tab:comparison-methods}.

\subsection{Forward model and variational methods}
Here, we provide the mathematical description of the signal formation and the deep image prior (DIP) reconstruction, which will serve as a basis for our method's derivation.
Let us represent the discretized expected image of the studied object as the vector $x^* \in \mathbb{R}^N$, where $N$ is the number of its pixels. We probe $x^*$ with a set of structured illumination patterns $\mathcal{W} = \{ w_1, w_2, \ldots, w_m\}$, which form the acquisition matrix $W = [ w_1, w_2, \ldots, w_m ] \in \mathbb{R}^{M \times N}$, where $M$ is the number of measurements. If we assume that the interaction between the property represented by $x^*$ and the illumination patterns $W$ is linear, then the recorded signal $y \in \mathbb{R}^M$, corrupted by an uncorrelated zero-mean noise $\epsilon$ is given by:
\begin{equation}
    \label{eq:forward_model}
    y = Wx^* + \epsilon = b + \epsilon
\end{equation}
where $b \in \mathbb{R}^M$ is the non corrupted signal.
The reconstruction aims to recover the vector $\hat{x}$ that is the most probable estimate of the clean image $x^*$. Regularized variational methods seek $\hat{x}$ by solving the following minimization problem:
\begin{equation}
    \label{eq:minimization_problem}
    \hat{x} = \argmin_x \frac{1}{2} \lVert Wx - y \rVert_2^2 + \lambda R(x)
\end{equation}
Where $R: \mathbb{R}^N \rightarrow \mathbb{R}$ denotes a regularization term that imposes some \textit{prior knowledge} on the reconstruction. This prior knowledge is often related to the structural properties of the to-be-reconstructed image, and it is supposed to enhance its signal-to-noise ratio. A common choice of regularization is the so-called total variation (TV) minimization, which promotes reconstructed images with sparse gradients. This, in turn, imposes short-range correlations in the reconstructions.
\\
Unsupervised generative machine learning methods have taken inspiration from \cref{eq:minimization_problem}~\cite{Habring2024}, and developed an image recovery scheme that seeks to solve a very similar minimization problem, but on the output of a machine-learning model (typically a CNN), as follows:
\begin{subequations}
    \label{eq:dip}
    \begin{flalign}
    \hat{\theta} & = \argmin_\theta \frac{1}{2} \lVert W N_{\theta}(r) - y \rVert_2^2 + \lambda R(N_{\theta}(r)) \label{eq:dip:learning} \\
    \hat{x} & = N_{\hat{\theta}}(r)
    \end{flalign}
\end{subequations}
where $r \in \mathbb{R}^N$ is any input image, which could be randomly chosen, $N_{\theta}: \mathbb{R}^N \rightarrow \mathbb{R}^N$ the model, and $\theta \in \mathbb{R}^T$ is its parameters vector with $T$ different parameters. In~\cite{Wang2022}, $r = W^\dagger y$ where $W^\dagger$ is the pseudo-inverse of $W$, making $r$ the least-squares reconstruction. The model $N_{\theta}$ learns to produce the image $\hat{x}$ that satisfies both the forward model \cref{eq:forward_model} and the regularization $R$, weighted by the $\lambda$ parameter. Neural networks have notions of non-locality and non-linearity in their response, which can behave as an additional prior on top of $R$~\cite{Lempitsky2018} when the training process is stopped before convergence. The $\lambda$ parameter is commonly determined through cross-validation, by creating a small leave-out set of measurements, and choosing the $\lambda$ that makes the model fit them best~\cite{Manni2023}.
\\
While \cref{eq:dip} has proven to help with the reconstruction of under-determined acquisitions~\cite{Wang2022}, it is not tuned to deal with strong random noise. Following the same derivation steps as in~\cite{Hendriksen2020}, the expected prediction error of the data fitting term in \cref{eq:dip:learning} is:
\begin{equation}
    \label{eq:exp_dip_error}
    \mathbb{E}_{y} \lVert W N_{\theta}(r) - y \rVert_2^2 = \mathbb{E}_{b, \epsilon} \lVert W N_{\theta}(W^\dagger(b + \epsilon)) - (b + \epsilon) \rVert_2^2
\end{equation}
which is intuitively minimized when $N_{\theta} = I$ (the identity matrix).
Thus, noise reduction is solely delegated to the regularization term $R$.
Therefore, this reconstruction technique might not be able to provide an advantage over traditional variational methods in this specific scenario. 

\subsection{Proposed method: Noise2Ghost (N2G)}
\begin{figure}[t]
    \centering
    \begin{minipage}[]{0.46\linewidth}
        \centering
        \begin{subfigure}[]{\linewidth}
            \includegraphics[width=\linewidth]{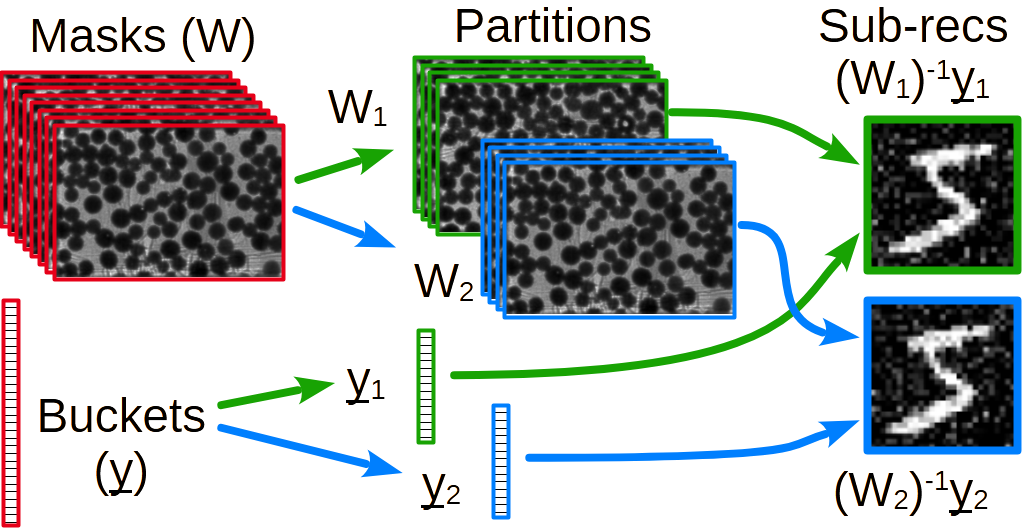}
            \subcaption{Realizations (masks and buckets) partitioning.}
            \label{fig:scheme:partitioning}
        \end{subfigure}
        \vskip\baselineskip 
        \setcounter{subfigure}{2}
        \begin{subfigure}[]{\linewidth}
            \includegraphics[width=\linewidth]{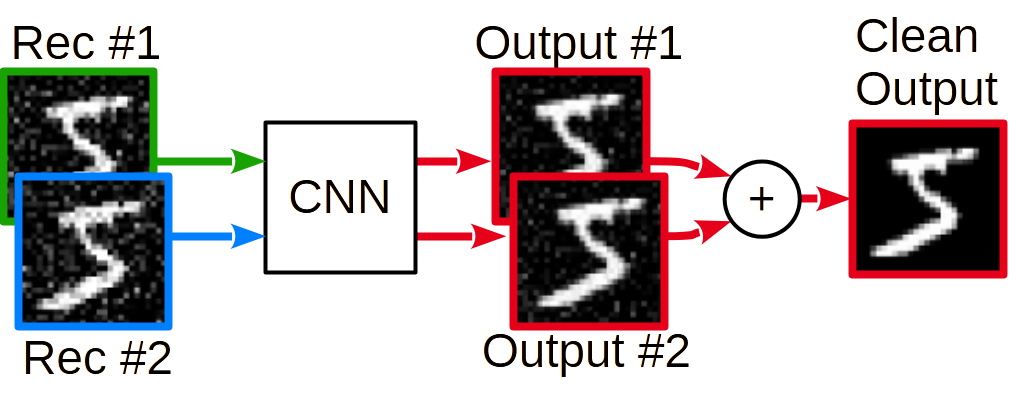}
            \subcaption{Prediction of the final reconstruction.}
            \label{fig:scheme:prediction}
        \end{subfigure}
    \end{minipage}
    \hfill
    \begin{minipage}[]{0.52\linewidth}
        \setcounter{subfigure}{1}
        \centering
        \begin{subfigure}[]{\linewidth}
            \includegraphics[width=\linewidth]{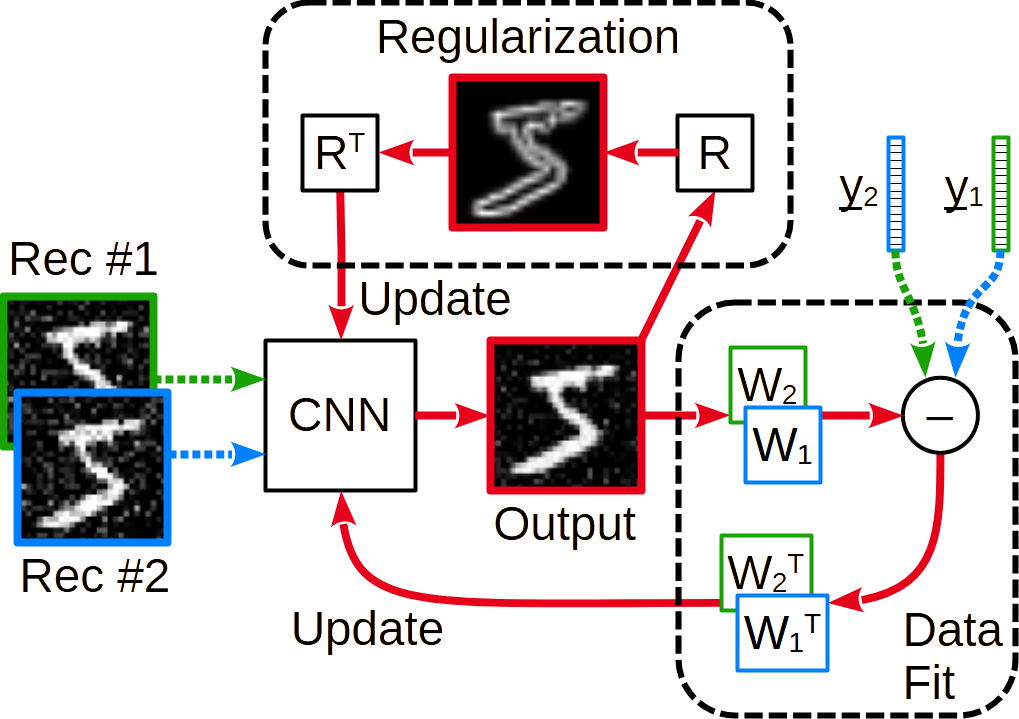}
            \subcaption{Self-supervised model (CNN) training.}
            \label{fig:scheme:training}
        \end{subfigure}
    \end{minipage}

    \caption{Schematic representation of the proposed method: (\textbf{\subref{fig:scheme:partitioning}})~The partitioning of the realizations set, generating the partial reconstructions (sub-recs),
    (\textbf{\subref{fig:scheme:training}})~the training procedure, and
    (\textbf{\subref{fig:scheme:prediction}})~the prediction of the final reconstruction.} 
    \label{fig:scheme}
    \vspace{-0.2cm}
\end{figure}
Here, we propose our CNN-based noise-tolerant unsupervised GI reconstruction method. It is inspired by DIP and N2N while addressing their limitations, namely accounting for the random noise in the optimization technique and the missing realization artifacts, respectively. The former is addressed in \cref{subsec:noise-reduction}, and the latter in \cref{subsec:N2I}.
We start from the assumption that by selecting a subset of the original realization set (i.e., the collection of all the acquired signals paired with the corresponding illumination patterns), we should obtain a degraded reconstruction due to a decrease in captured information. However, the degraded reconstruction should still represent the same signal as the reconstruction from the full realization set.
This means that if we partition the realizations into subsets with equal numbers of realizations, we should obtain equivalent but different and degraded sub-reconstructions.
\\
We partition the realizations in $k$ splits $k \in [1,K]$, producing $K$ different sub-reconstructions (i.e., a k-tuple of reconstructions).
Each sub-reconstruction in the said k-tuple has unique noise characteristics compared to the other reconstructions, and it provides different information on the FoV. From the partitioned data vectors $y_{k}$ we compute the sub-reconstructions $x_{k}$ as follows (\cref{fig:scheme:partitioning} and \cref{alg:split}):
\begin{equation}
    \label{eq:sub-realizations}
    x_{k} = W_{k}^{\dagger} y_{k} = W_{k}^{\dagger} (b_{k} + \epsilon_{k})
\end{equation} 
where $W_{k}^{\dagger}$ is pseudo-inverse of the partition forward model $W_{k}$, and $b_{k}$ is the clean data. The $x_{k}$ solutions can be reconstructed with a common least-squares solver.
\begin{algorithm}
    \caption{Split Realizations and Least-Squares Reconstructions} \label{alg:split}
    \begin{algorithmic}[1]
        \Require Realizations $W$, measurements $y$, number of splits $K$
        \Ensure Sub-reconstructions $\{x_k\}_{k=1}^K$
        \State $s \gets Shuffle(Range(1, M))$ \Comment{Randomly shuffle bucket indices [1, M]}
        \For{$k = 1$ to $K$}
            \State $s_k \gets s[M(k/K) + 1, M((k+1)/K)]$ \Comment{Slice $s$}
            
            \State $W_k, y_k \gets W[s_k], y[s_k]$ \Comment{Partition $W$ and $y$}
            
            \State $x_k \gets W_k^\dagger y_k$ \Comment{Least-squares reconstruction}
        \EndFor
        
        \State \Return $\{x_k\}_{k=1}^K$, $\{W_k, y_k\}_{k=1}^K$
    \end{algorithmic}
\end{algorithm}
\\
Each sub-reconstruction $x_{k}$ represents the same signal but has a different noise. In N2G, we feed each of these sub-reconstructions to the model, and then we optimize the following minimization problem (\cref{fig:scheme:training} and \cref{alg:train}) by comparing them against the realizations that were not used to reconstruct them:
\begin{subequations}
    \label{eq:n2g}
    \begin{flalign}
    \hat{\theta} & = \argmin_\theta \frac{1}{2} \sum_{k} \sum_{i \neq k}\lVert W_{i}N_\theta(x_{k}) - y_{i} \rVert_2^2 + \lambda R(N_\theta(x_{k})) \label{eq:n2g:learning}  \\
    \hat{x} & = \frac{1}{K}\sum_{k}N_{\hat{\theta}}(x_{k}). \label{eq:n2g:prediction}
    \end{flalign}
\end{subequations}
Since the different sub-reconstructions $x_{k}$ should represent the same signal, we would like the model to produce the same result for each. Thus, the model should learn to differentiate between the actual signal and random noise.
This is in opposition to \cref{eq:dip:learning}, where the same realizations were used for the input reconstruction and as targets in the learning loss.
Finally, \cref{fig:scheme:prediction} presents how \cref{eq:n2g:prediction} produces the reconstructed images
\begin{algorithm}
    \caption{N2G Training Loop} \label{alg:train}
    \begin{algorithmic}[1]
        \Require Sub-reconstructions $\{x_k\}_{k=1}^K$, realizations $\{W_k, y_k\}_{k=1}^K$, regularization $R$, parameter $\lambda$
        \Ensure Trained model parameters $\hat{\theta}$
        
        \State Initialize model $N_\theta$ randomly
        \For{each epoch}
            \For{$k = 1$ to $K$}
                \For{$i = 1$ to $K$, $i \neq k$}
                    \State $\mathcal{L}_{k,i} = \|W_i N_\theta(x_k) - y_i\|_2^2$ \Comment{Self-supervised losses}
                \EndFor
                \State $\mathcal{L}_{\text{TV},k} = R(N_\theta(x_k))$ \Comment{Prior (TV) losses}
                \State $\mathcal{L}_k \gets (\frac{1}{2} \sum_{i \neq k} \mathcal{L}_{k,i}) + \lambda \mathcal{L}_{\text{TV},k}$  \Comment{Sub-reconstruction losses}
            \EndFor
            \State $\theta \gets \theta - \nabla_\theta (\sum_k \mathcal{L}_k)$ \Comment{Update parameters}
        \EndFor
        
        \State \Return $\hat{\theta}$
    \end{algorithmic}
\end{algorithm}

\subsection{Self-supervised random noise reduction}
\label{subsec:noise-reduction}
We now derive the noise-reduction properties of \cref{eq:n2g:learning}.
Reconstruction noise comes from two main sources: Measurement noise; and the lack of information from a low number of realizations, which creates the so-called \textit{missing realization} artifacts. These noise sources can be represented in the sub-reconstructions $x_{k}$ as follows:
\begin{subequations}
    \label{eq:missing_realizations}
    \begin{flalign}
&    x_{k} = x + v_{k} + W_{k}^{\dagger}\epsilon_{k} = x + v_{k} + t_{k} \label{eq:sub-rec} \\
&    W_{k} v_{k} = 0 \\
&    W_{k} x_{k} = y_{k} + \epsilon_{k}
    \end{flalign}
\end{subequations}
where $W_{k}^{\dagger}\epsilon_{k} = t_{k}$ is the pseudo-inverse of the bucket measurement noise, while the vector $v_{k} \subseteq \ker W_{k}$ describes spurious solutions that could be added to the reconstruction, and still fit the data.
Our method focuses on the random measurement noise, and the desired noise reduction effect is given by the specific construct of the data fitting term from \cref{eq:n2g:learning}. When projecting the predicted image onto the target data from the other splits, the projected noise from the $k1$ split will be inconsistent with the random noise from the other $K - 1$ splits.
This can be seen by substituting \cref{eq:sub-rec} and \cref{eq:sub-realizations} into the data fitting term of \cref{eq:n2g:learning}:
\begin{equation}
    \label{eq:minimization_results}
    \frac{1}{2}\sum_{i\neq k}\lVert W_{i}N_\theta(x + v_{k} + W^{\dagger} \epsilon_{k}) - (b + \epsilon_{i}) \rVert_2^2
\end{equation}
As $\epsilon_{k}$ is independent of any other $\epsilon_i$ and their domains do not overlap (Appendix A from~\cite{Hendriksen2020}), the expected prediction error of \cref{eq:minimization_results} becomes:
\begin{subequations}
    \label{eq:supervised_error}
    \begin{flalign}
    \mathbb{E}_{b,\epsilon,i\neq k} & \lVert W_{i} N_\theta(x + v_{k} + W^{\dagger} \epsilon_{k}) - b_i \rVert_2^2 + \mathbb{E}_{b,\epsilon,i\neq k} \lVert b_i - (b_i - \epsilon_{i}) \rVert_2^2 \\
    = \mathbb{E}_{b,i\neq k} & \lVert W_{i} N_\theta(x_{k}) - b_i \rVert_2^2 + \mathbb{E}_{\epsilon,i\neq k} \lVert \epsilon_{i} \rVert_2^2
    \end{flalign}
\end{subequations}
where the term $\lVert W_{i} N_\theta(x_{k}) - b_i \rVert_2^2$ is the \textit{supervised reconstruction error}, and the term $\lVert \epsilon_{i} \rVert_2^2$ is the signal noise variance.
In the supervised reconstruction term, the acquisition noise is not present. This means that it is equivalent to having noiseless realizations as the targets of our N2G reconstruction, thus providing the noise suppression characteristics of Noise2Ghost.
The signal noise variance is just a constant in the objective function, which only depends on the acquisition noise and should not contribute to the objective function gradient. In reality, it may introduce some noise in the optimization gradients at very high noise levels. This could interfere with the reconstruction, guiding it towards local minima. As for N2N and derived methods, this often results in excessive blur and resolution loss in the reconstructions.

\subsection{Relationship with Noise2Inverse}
\label{subsec:N2I}
The Noise2Inverse method addresses the problem in \cref{eq:n2g} directly in the reconstruction domain~\cite{Hendriksen2020}, by solving the following problem:
\begin{subequations}
    \label{eq:n2i}
    \begin{flalign}
    \hat{\theta} & = \argmin_\theta{\frac{1}{2} \sum_{k}\lVert N_{\theta}(x_{k}) - \frac{1}{K - 1}\sum_{i \neq k}(x_i) \rVert_2^2} \label{eq:n2i:learning}  \\
    \hat{x} & = \frac{1}{K}\sum_{k}N_{\hat{\theta}}(x_{k}). \label{eq:n2i:prediction}
    \end{flalign}
\end{subequations}
where $x_i = W_{i}^{\dagger} y_i$ is the pseudo-inverse (LS reconstruction) of the partial set of masks $W_i$ and corresponding buckets $y_i$ with $i \in [1, K]$.
Theoretical grounds support this formulation, and it should retrieve the same result as \cref{eq:n2g}, regarding its noise reduction aspect. However, it does not address the missing realization artifacts and suffers from strong technical limitations.
In ghost imaging, we aim at reducing the number of realizations as much as possible, which results in faster acquisition time and potentially reduced dose deposition. This exacerbates the rise of related artifacts. Missing realization and random noise artifacts are strongly structured and have long-range correlations, i.e., they generally cover the whole FoV.
Commonly used models like CNNs have a limited receptive field (i.e., the region around each pixel that the CNN probes and that provides the context for the said pixel) that may not extend to the entire FoV. Depending on the model, this could have a strong negative impact on the denoising performance.
\\
More precisely, about the missing realization artifacts, in~\cite{Hendriksen2020} it is stated that~\cref{eq:n2i} is unable to cope with them. The LS sub-reconstructions cope with the missing realizations by acting as if their bucket values were set to 0. This selects one specific solution $v_k$ for each $x_k$, out of the infinite solutions in $\ker W_k$ that satisfy \cref{eq:forward_model}. This specific choice of $v_k$ is probably incorrect, but is imposed by N2I as the learning target in \cref{eq:n2i}. The model learns to discard the inconsistent features of different $v_k$ functions, selecting a specific $v \in \ker W$ as a common ground.
\\
The formulation we propose in (\cref{eq:n2g}) circumvents this limitation by using the measured buckets as ``target''. Thus, N2G does not embed any specific $v_k$ into the learning target. This allows the model to select a $v \in \ker W$ that satisfies both the measured buckets and the prior knowledge injected by the regularization term $R$ and the regularization properties of CNNs~\cite{Lempitsky2018}.
\\
A possible approach to overcome the just-discussed limitations of \cref{eq:n2i} would be to use variational methods (like in \cref{eq:minimization_problem}) to compute the sub-reconstructions. These non-linear methods are known to remove or strongly reduce these long-range artifacts. However, as discussed in~\cite{Hendriksen2020}, their non-linear nature contrasts with the assumptions of \cref{eq:n2i}. This may then result in a degraded reconstruction quality.

\subsection{Data augmentation}
Thanks to the assumption that any permutation (reordering) of the acquired realizations provides the same reconstruction, we can achieve simple and effective data augmentation: We can increase the number of training examples for the self-supervised method from the measured data.
We permute our dataset in $p$ arrangements with $p \in [1, P]$ and partition each arrangement in $k$ splits $k \in [1,K]$, producing $KP$ different sub-reconstructions (i.e., $P$ k-tuples of reconstructions).
In practice, the permutations $P$ provide unique mixtures of missing realization artifacts and random noise, while the $K$ splits are still the underlying mechanism for noise reduction.
This modifies \cref{eq:n2g} into:
\begin{subequations}
    \label{eq:n2g-augm}
    \begin{flalign}
    \hat{\theta} & = \argmin_\theta{\frac{1}{2} \sum_{p,k}\lVert WN_{\theta}(x_{p,k}) - y \rVert_2^2 + \lambda R(N_\theta(x_{p,k}))} \label{eq:n2g-augm:learning}  \\
    \hat{x} & = \frac{1}{PK}\sum_{p,k}N_{\hat{\theta}}(x_{p,k}).
    \end{flalign}
\end{subequations}

\subsection{Models}
Here, we briefly discuss the machine learning models, in particular neural networks (NN), that N2G could leverage.
The model should accept multiple images of the same object as input. This is incompatible with INRs, which learn coordinate-based representations of the signal and thus take coordinates as inputs. CNNs and other MLP-based NNs are instead good candidates. Both NN types take images as input and can learn short- and long-range pixel correlations.
Thus, N2G is not tied to a specific NN architecture and can take advantage of future developments in that area.
MLP-based NNs examples include the Visual Transformer~(ViT)~\cite{Dosovitskiy2021} and the swin-transformer~\cite{Liu2021}. CNNs examples include U-Net~\cite{Ronneberger2015}, DnCNN~\cite{Zhang2017}, and MS-D~net~\cite{Pelt2018}.
Regarding the model size, one should try to work with small NNs to minimize the computation time and resources and to limit overfitting. For instance, we suggest models with parameter numbers in the order of 100k for a $100 \times 100$ pixels image. However, some models like MS-D net could achieve similar results with 10-20k parameters for similar image sizes. In our tests, we had the best results with U-Net and DnCNN with around 100-300k parameters. Further details will be provided with the results.

\subsection{Computational Performance \& Requirements}
N2G presents specific computational and memory demands that require to be considered. It requires a $KP$-fold increase in operations, compared to GIDC (e.g., $\times 24$ for the tests performed in \cref{sec:results}). This does not correspond to a proportional increase in computational time, thanks to batching and GPU parallelism, as detailed in the Supplementary Information, Sec. 5. The actual increase in reconstruction time is approximately 2.1 to 2.5 times. On the other hand, memory usage increases proportionally, requiring $\times KP$ activation weights during backpropagation. This is critical for GPUs, where memory constraints are typically more stringent compared to CPU RAM.
The increased memory requirements can limit the size of images and the extent of noise reduction that can be processed with N2G. To mitigate this, gradient accumulation can be used, where backpropagation weights are accumulated over smaller batches of model passes. While this approach can impact reconstruction performance, it effectively allows for the processing of larger reconstruction sizes. Looking ahead, this strategy could enable implementing multi-GPU reconstructions, further enhancing the capabilities of N2G.
Furthermore, adjustments of ``activation checkpointing'' could further yield reductions in GPU memory consumption.
\\
Note that, due to the intrinsic iterative nature of the methods discussed here, their use in real-time applications should be considered relative to the acquisition time. As presented in Sec. 5 of the Supplementary Information, reconstructions can take from several seconds to several minutes. This is incompatible with high-throughput acquisitions, where full datasets are produced short time-frames of less than a second or millisecond. However, given the sequential nature of GI acquisitions, this scenario is unlikely. In particular, for x-ray GI, the current acquisition time is on the order of a few hours per image~\cite{Manni2023}. In this context, N2G's increased computational time requirements over the existing methods are negligible, even when performing a full hyperparameter grid search.

\section{Results}
\label{sec:results}
We now analyze N2G's performance against existing unsupervised methods for reconstructing both synthetic and real data. The reference methods are LS, TV, the GIDC algorithm from~\cite{Wang2022} (based on DIP), and INRs. For the synthetic test cases, we first derive a method to module the noise and compare different noise levels. Additionally, for the synthetic study cases, we also compare the results against N2V and N2I, two common self-supervised denoising methods.

\subsection{Synthetic data generation \& quantification}
We first evaluate the performance of the proposed method on synthetic data, as it provides the ground truth and precisely allows us to control the noise added to the data. In Sec. 6 of the Supplementary Information, we present and analyze additional synthetic test cases, with increasing image complexity.
\\
We extracted a $100 \times 100$ pixels image to be our ground truth (phantom, e.g., from \cref{fig:chrom-cmprss:ph,fig:chrom-noise:ph}) from~\cite{Tseng2023}, a curated collection of chromosome images. We generated synthetic random masks with a normalized half-Gaussian distribution and forward projected the phantom to create the corresponding buckets.
We added Poisson noise to the buckets by first multiplying them by the maximum expected emitted number of photons per pixel. We fed the resulting mean expected photons per realizations to the Poisson distribution function. We then normalized the signals by dividing the noisy buckets by the same number of expected photons, resulting in signals:
\begin{equation}
    \label{eq:sym-noise-model}
    y_m = \mathcal{P}\left( C \textstyle \sum_n^N w_{mn} x_n \right) / C.
\end{equation}
This expression is derived in Sec. 2 of the Supplementary Information for the specific case of XRF imaging.
We use the constant $C$ to modulate the noise and evaluate its effect on the reconstruction performance of the proposed method.
The advantage of this noise model is that it gives emphasis to the number of photons delivered to the sample.
\\
We evaluate the reconstruction algorithms both visually and with multiple objective metrics: the mean square error (MSE), the peak signal-to-noise ratio (PSNR), the structural similarity index (SSIM), and the resolution, which is estimated from the Fourier ring correlation (FRC). Their formulas are provided in Sec. 1 of the Supplementary Information, while here we shortly describe their meaning. The MSE evaluates the average pixel-wise difference between the phantom and the reconstruction intensity values. The SSIM calculates the similarity between these two images by comparing the inter-dependencies of spatially neighboring pixels over various window sizes~\cite{Wang2004}. PSNR is based on MSE, but it focuses on the peak intensity power instead of its average.
Finally, the FRC computes the similarity in Fourier space between the phantom and the reconstruction for all the sampled frequencies. We estimate the resolution of the reconstruction by finding the intersection of the resulting curve with the threshold function defined in~\cite{VanHeel2005}.

\subsection{Synthetic data reconstructions across various noise levels}
\begin{figure}
    \centering
    \includegraphics[width=\linewidth]{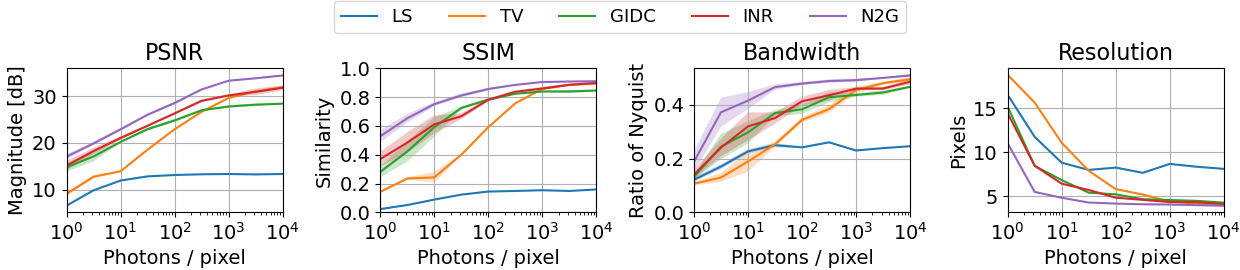}

    \caption{Synthetic data GI reconstructions comparison (chromosomes phantom), for 5$\times$ compression ratio, and maximum emitted photons per pixel per realization in the range [$10^0$, $10^4$]. We compare peak signal-to-noise ratio (PSNR), structural similarity index (SSIM), and bandwidths (Nyquist frequency) against the ground truth. Higher values indicate higher reconstruction quality. We also show the corresponding image resolutions in pixels.}
    \label{fig:chrom-progr}
    \vspace{-0.2cm}
\end{figure}
We present three different study cases: In the first case (\cref{fig:chrom-progr}), we present the performance of the above-mentioned different algorithms (LS, TV, GIDC, INR, and N2G) across multiple noise levels (i.e., photons per pixel) using only 2000 realizations, corresponding to 20\% of the total number of pixels.
The simulated maximum emitted photons per pixel per realization (i.e., constant $C$) intensities span the range of $[10^0, 10^4]$. The lower margin means that each pixel is illuminated on average by around one photon. The upper margin was empirically chosen well above the threshold where noise is not more the limiting factor of the reconstruction quality (i.e., it becomes the missing realizations).
To add some statistical significance to our tests, we generated five different sets of realizations for each noise level. We then independently reconstruct these five sets and compute their individual evaluation metrics (e.g., mean error against the phantom, etc). The presented results report the average and standard deviation of the said computed metrics for each noise level.
\\
For the TV, GIDC, INR, and N2G reconstructions, the regularization weight $\lambda$ was chosen by cross-validation, by putting aside a set of realizations ($10 \%$ of the total) and using them to assess the best value over a wide range~\cite{Manni2023}. For TV, we use a $\lambda \in [10^{-5}, 10^{-1}]$, while for GIDC, INR, and N2G we used a $\lambda \in [10^{-7}, 10^{-4}]$. This difference in range is due to the different data normalization used by the different algorithm implementations.
As regularization $R$ for GIDC, INR and N2G, we also used the TV term. Although stronger and more advanced priors could have been used, we wanted a direct comparison against the well-known and used TV-min-based variational regularization. Moreover, the TV term has very simple and widely understood implementation and behavior.
To improve the procedure's robustness, we ran it three times with different cross-validation sets for each run and averaged the cross-validation loss values between those different runs.
For the INR reconstructions, we used a 3-layer feed-forward NN with 512 neurons per layer, sinusoidal activation functions~\cite{Sitzmann2019}, and 256 Fourier embeddings~\cite{Tancik2020}.
Regarding specifically GIDC and N2G, we used a U-net~\cite{Ronneberger2015} with 20 features and 3 decomposition levels ($\sim$ 192k parameters). We learned the U-net and INR weights with 5000 and 7000 epochs of the Adam optimizer~\cite{Kingma2015} respectively, with a learning rate of $3\cdot10^{-4}$ and weight decay of $1 \cdot 10^{-2}$ for both architectures. We selected the model weights corresponding to the epoch in which the cross-validation loss function is minimized. For N2G, we used 4 partitions and 6 permutations ($K = 4$, and $P = 6$). For a more detailed discussion of the choice of the $K$ and $P$ parameters, we refer to Sec. 4 of the Supplementary Information.
\begin{figure}
    \centering
    \begin{subfigure}[b]{0.8\linewidth}
        \includegraphics[width=\linewidth]{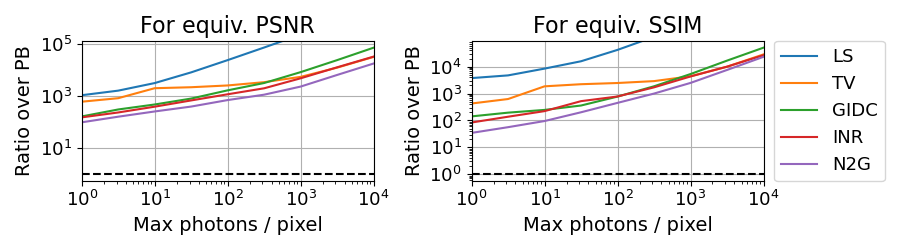}
        \subcaption{Total FoV dose (against PB)}
        \label{fig:chrom-progr-dose:tot}
    \end{subfigure}
    \vskip\baselineskip 
    \begin{subfigure}[b]{0.8\linewidth}
        \includegraphics[width=\linewidth]{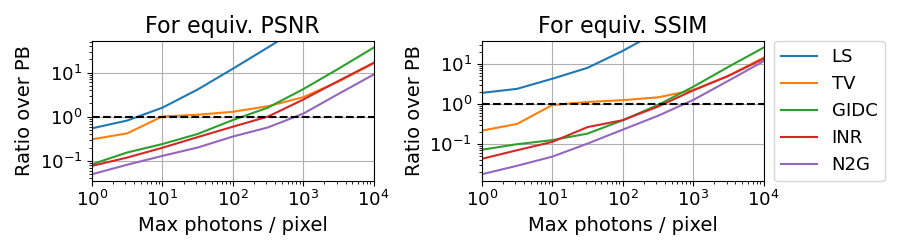}
        \subcaption{Average maximum pixel dose / unit-time (against PB)}
        \label{fig:chrom-progr-dose:avg-max}
    \end{subfigure}
    \vskip\baselineskip 
    \begin{subfigure}[b]{0.8\linewidth}
        \includegraphics[width=\linewidth]{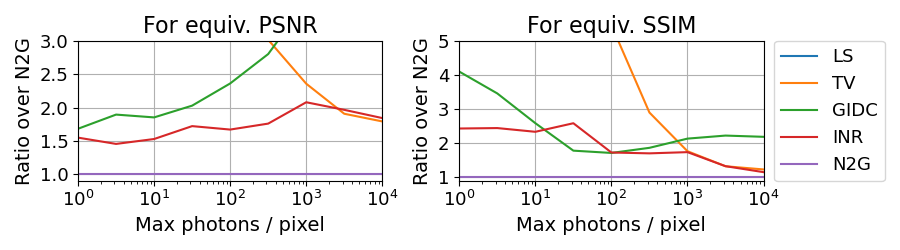}
        \subcaption{Average maximum pixel dose / unit-time (against N2G)}
        \label{fig:chrom-progr-dose:avg-max-n2g}
    \end{subfigure}

    \caption{Dose requirements for the chromosomes phantom, with 5$\times$ compression ratio, and varying noise levels: (\textbf{\subref{fig:chrom-progr-dose:tot}})~total required dose for each algorithm to obtain equivalent PSNR and SSIM as a PB acquisition; (\textbf{\subref{fig:chrom-progr-dose:avg-max}})~same plot as (\textbf{\subref{fig:chrom-progr-dose:tot}})~for the average maximum pixel dose per illumination; and (\textbf{\subref{fig:chrom-progr-dose:avg-max-n2g}})~for N2G.}
    \label{fig:chrom-progr-dose}
    \vspace{-0.2cm}
\end{figure}
\\
On the top left side of~\cref{fig:chrom-progr}, we present the peak signal-to-noise ratio (PSNR) of the reconstructions against the phantom.
It is evident that N2G consistently achieves superior reconstructions, with PSNRs surpassing those of competing methods by 1–2 dB, highlighting its robustness across diverse imaging conditions. Nonetheless, its advantage seems to diminish under low-photon scenarios. While counterintuitive, this is easily explained by the fact that the lower frequencies are less affected by Poisson noise. PSNR takes all frequencies into account, resulting in a bias towards lower frequencies with low photon counts. The SSIM, by construction, is instead less affected by lower frequencies. For this reason, on the top right side of~\cref{fig:chrom-progr}, we see that N2G has a growing lead in SSIM for lower photon counts.
We also observe that for high photon numbers (equating to low noise), the compression ratio becomes the limiting factor for reconstruction accuracy, and both PSNR and SSIM reach a plateau.
From~\cref{fig:chrom-progr}, we have instead a perspective on how the resolution is degraded with increasing noise levels. We observe that N2G better preserves the reconstruction resolution, by showing a drop-off at lower photon counts than other methods.
\\
Finally, we explore dose considerations by comparing GI with PB and assess N2G's performance against LS, TV, GIDC, and INR.
In particular, we evaluate the total sample and average maximum pixel dose necessary to achieve a GI image with PSNR or SSIM quality comparable to a PB scan.
From figure \cref{fig:chrom-progr-dose:tot}, we observe that in terms of total dose, PB always outperforms GI in the chosen compression ratio, regardless of the method used, although N2G narrows the difference.  
Concerning the average maximum pixel dose per exposure (\cref{fig:chrom-progr-dose:avg-max}), GI generally exceeds PB's performance when the main noise source is detection noise (except when LS is used). However, N2G still decreases dose requirements compared to the other GI reconstruction techniques and widens the region where GI is preferable over PB. \Cref{fig:chrom-progr-dose:avg-max-n2g} clearly shows this trend by plotting the required dose ratio from LS, TV, GIDC, and INR to obtain the same performance as N2G.

\subsection{In-depth analysis of noisy reconstructions}
\begin{figure}[ht]
    \raggedright
    \begin{subfigure}[b]{0.1166\linewidth}
        \includegraphics[width=\linewidth]{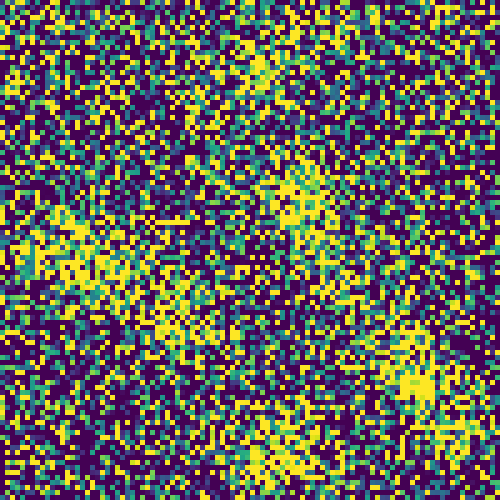}
        \subcaption{LS}
        \label{fig:chrom-cmprss:ls}
    \end{subfigure}
    \begin{subfigure}[b]{0.1166\linewidth}
        \includegraphics[width=\linewidth]{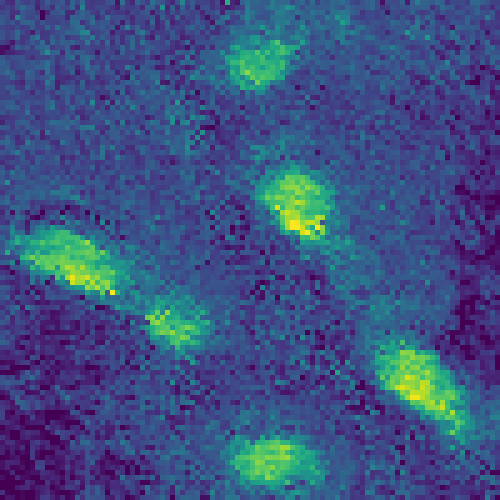}
        \subcaption{N2V}
        \label{fig:chrom-cmprss:n2v}
    \end{subfigure}
    \begin{subfigure}[b]{0.1166\linewidth}
        \includegraphics[width=\linewidth]{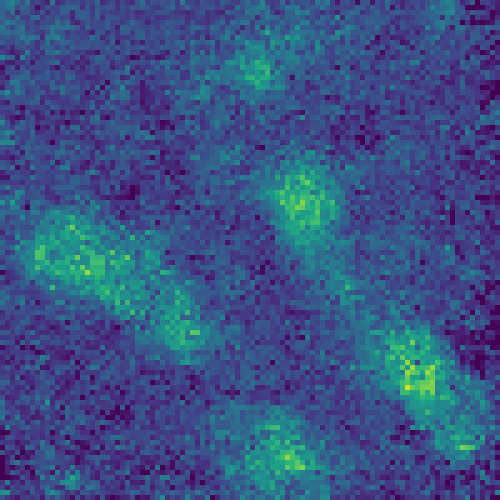}
        \subcaption{N2I}
        \label{fig:chrom-cmprss:n2i}
    \end{subfigure}
    \begin{subfigure}[b]{0.1166\linewidth}
        \includegraphics[width=\linewidth]{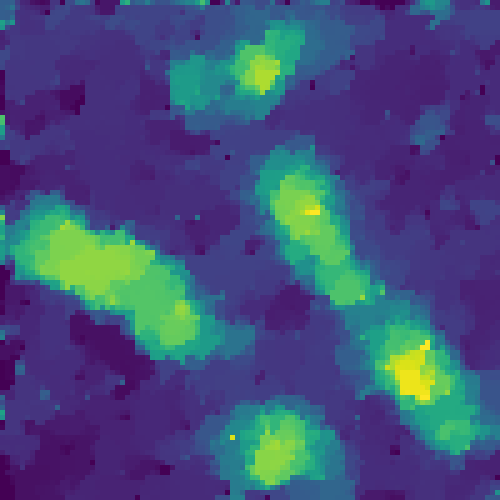}
        \subcaption{TV}
        \label{fig:chrom-cmprss:tv}
    \end{subfigure}
    \begin{subfigure}[b]{0.1166\linewidth}
        \includegraphics[width=\linewidth]{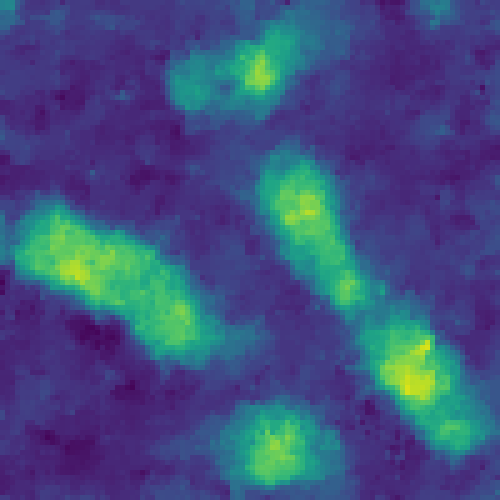}
        \subcaption{GIDC}
        \label{fig:chrom-cmprss:gidc}
    \end{subfigure}
    \begin{subfigure}[b]{0.1166\linewidth}
        \includegraphics[width=\linewidth]{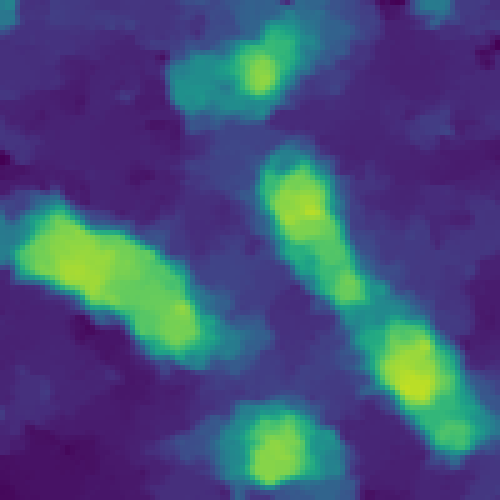}
        \subcaption{INR}
        \label{fig:chrom-cmprss:inr}
    \end{subfigure}
    \begin{subfigure}[b]{0.1166\linewidth}
        \includegraphics[width=\linewidth]{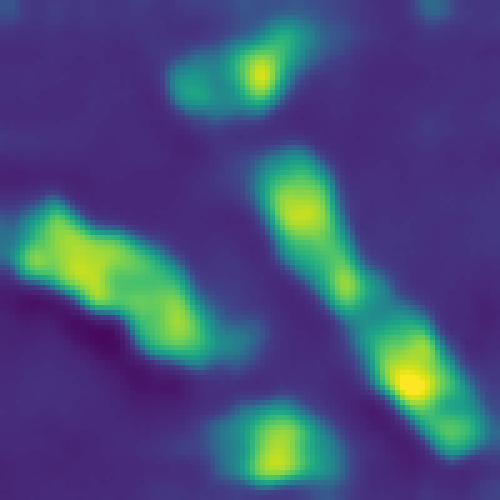}
        \subcaption{N2G}
        \label{fig:chrom-cmprss:n2g}
    \end{subfigure}
    \begin{subfigure}[b]{0.1439\linewidth}
        \includegraphics[width=\linewidth]{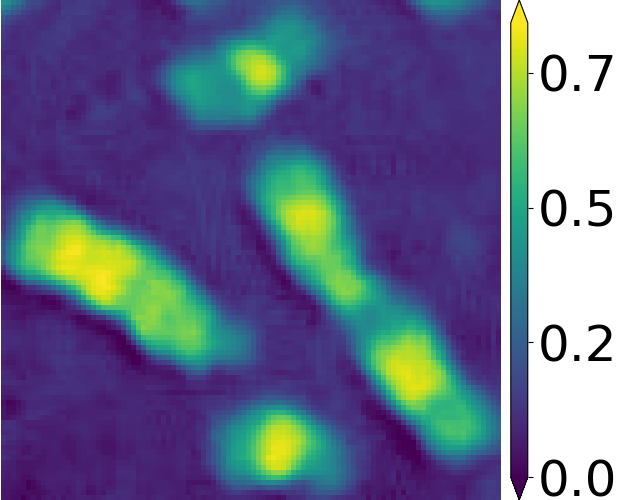}
        \subcaption{Phantom}
        \label{fig:chrom-cmprss:ph}
    \end{subfigure}
    \vskip\baselineskip 
    \centering
    \begin{subfigure}[b]{0.65\linewidth}
        \includegraphics[width=\linewidth]{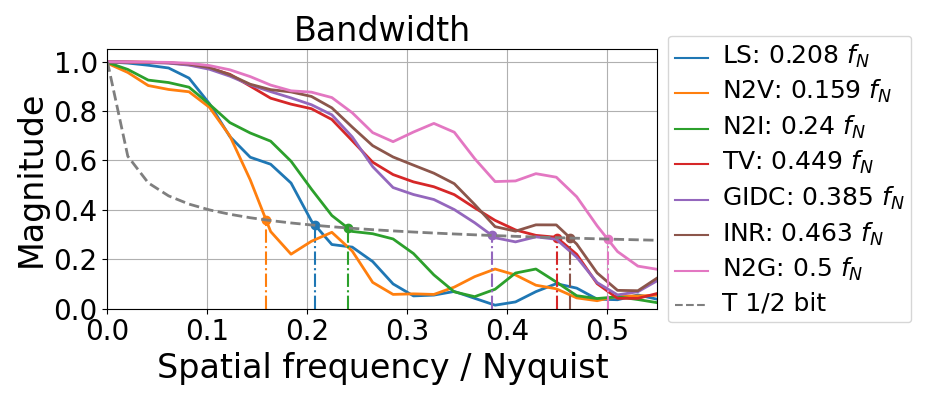}
        \subcaption{FRC}
        \label{fig:chrom-cmprss:frc}
    \end{subfigure}
    \vskip\baselineskip 
    \begin{subfigure}[b]{0.70\linewidth}
    \begin{center}
        \setlength{\tabcolsep}{3pt} 
        \footnotesize
        \begin{tabular}{|r||c|c|c|c|c|c|c|}
            \hline
             & LS & N2V & N2I & TV & GIDC & INR & N2G \\
            \hline \hline
            MSE $\downarrow$ & 0.400 & 0.021 & 0.021 & 0.004 & 0.004 & 0.002 & \textbf{0.001} \\
            \hline
            PSNR $\uparrow$ & 2.50 & 15.30 & 15.27 & 22.44 & 22.70 & 25.00 & \textbf{26.89} \\
            \hline
            SSIM $\uparrow$ & 0.030 & 0.235 & 0.246 & 0.656 & 0.672 & 0.756 & \textbf{0.827} \\
            \hline
            Resolution $\downarrow$ & 9.6 & 12.6 & 8.3 & 4.4 & 5.2 & 4.3 & \textbf{4.0} \\
            \hline
        \end{tabular}
        \subcaption{Performance comparison.}
        \label{fig:chrom-cmprss:table}
    \end{center}
    \end{subfigure}

    \caption{GI reconstructions of the chromosomes phantom, with 10$\times$ compression ratio and moderate Poisson noise: Mean noise fluctuations $\sim 24.5\%$ of the mean clean bucket fluctuations. From~(\textbf{\subref{fig:chrom-cmprss:ls}}) to~(\textbf{\subref{fig:chrom-cmprss:n2g}}) we show the reconstructions of~(\subref{fig:chrom-cmprss:ph}) with LS, TV, GIDC, INR and N2G respectively. In~(\textbf{\subref{fig:chrom-cmprss:frc}}) and in~(\textbf{\subref{fig:chrom-cmprss:table}}) we present the Fourier ring correlation against~(\textbf{\subref{fig:chrom-cmprss:ph}}) and various performance metrics of each reconstruction, respectively.}
    \label{fig:chrom-cmprss}
    \vspace{-0.2cm}
\end{figure}
We now provide an in-depth analysis of two examples: (a) high compression ratio and moderate noise and (b) low compression ratio and very high noise (higher than signal).
\\
We present the results for (a) in~\cref{fig:chrom-cmprss}. We produced 1000 realizations, resulting in a 10$\times$ compression ratio. The applied Poisson noise of $10^2$ maximum emitted photons per pixel per realization (i.e., constant $C = 10^2$) resulted in fluctuations equal to $24.5\%$ of the mean clean bucket fluctuations. Comparing reconstructions from~\crefrange{fig:chrom-cmprss:ls}{fig:chrom-cmprss:n2g} against~\cref{fig:chrom-cmprss:ph}, we note that except for the LS reconstruction, all other algorithms, including N2G, can correctly reconstruct the shape of the phantom. The only visible difference is that N2G's reconstruction presents much less high-frequency noise.
This results in the N2G reconstruction having the highest resolution (in pixels), PSNR, and SSIM values, and the lowest MSE among all reconstructions. These results are summarized in~\cref{fig:chrom-cmprss:table}.
\begin{figure}[ht]
    \raggedright
    \begin{subfigure}[b]{0.1166\linewidth}
        \includegraphics[width=\linewidth]{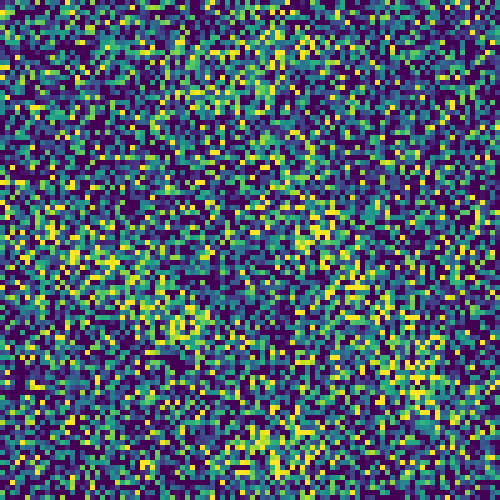}
        \subcaption{LS}
        \label{fig:chrom-noise:ls}
    \end{subfigure}
    \begin{subfigure}[b]{0.1166\linewidth}
        \includegraphics[width=\linewidth]{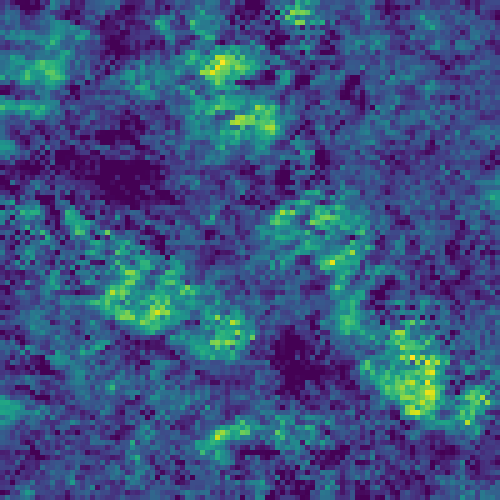}
        \subcaption{N2V}
        \label{fig:chrom-noise:n2v}
    \end{subfigure}
    \begin{subfigure}[b]{0.1166\linewidth}
        \includegraphics[width=\linewidth]{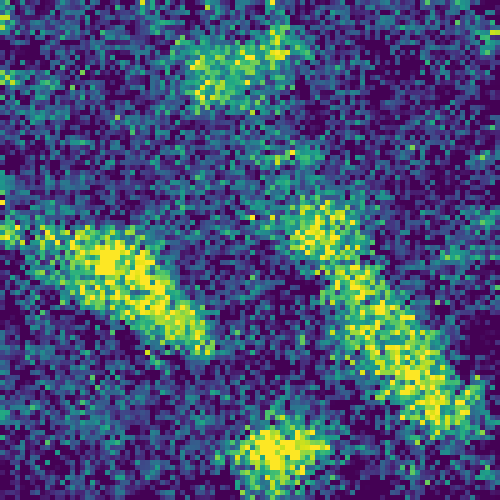}
        \subcaption{N2I}
        \label{fig:chrom-noise:n2i}
    \end{subfigure}
    \begin{subfigure}[b]{0.1166\linewidth}
        \includegraphics[width=\linewidth]{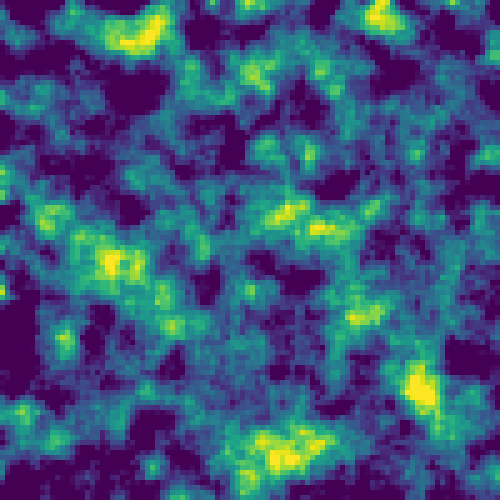}
        \subcaption{TV}
        \label{fig:chrom-noise:tv}
    \end{subfigure}
    \begin{subfigure}[b]{0.1166\linewidth}
        \includegraphics[width=\linewidth]{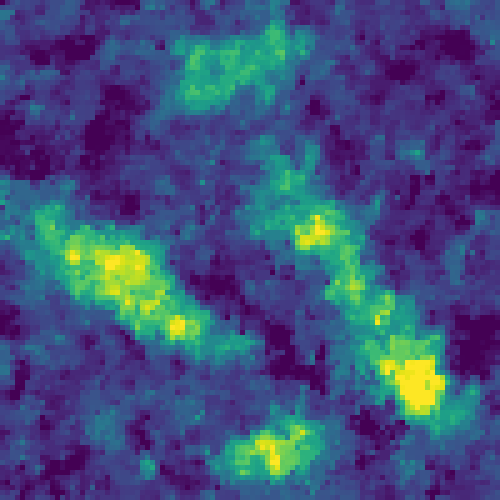}
        \subcaption{GIDC}
        \label{fig:chrom-noise:gidc}
    \end{subfigure}
    \begin{subfigure}[b]{0.1166\linewidth}
        \includegraphics[width=\linewidth]{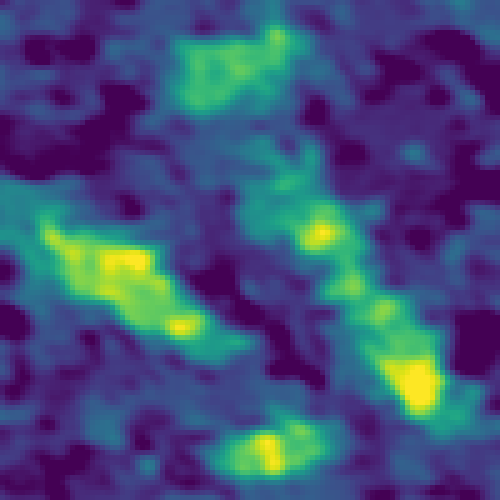}
        \subcaption{INR}
        \label{fig:chrom-noise:inr}
    \end{subfigure}
    \begin{subfigure}[b]{0.1166\linewidth}
        \includegraphics[width=\linewidth]{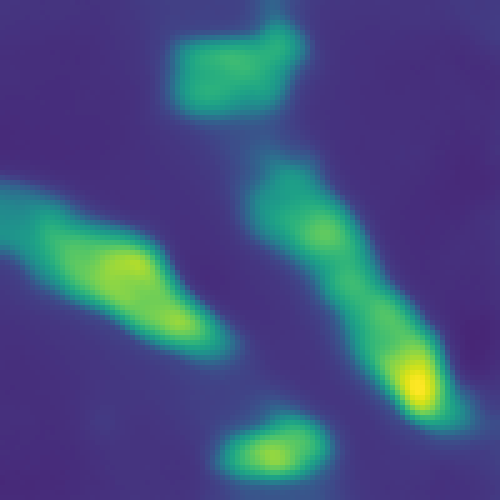}
        \subcaption{N2G}
        \label{fig:chrom-noise:n2g}
    \end{subfigure}
    \begin{subfigure}[b]{0.1439\linewidth}
        \includegraphics[width=\linewidth]{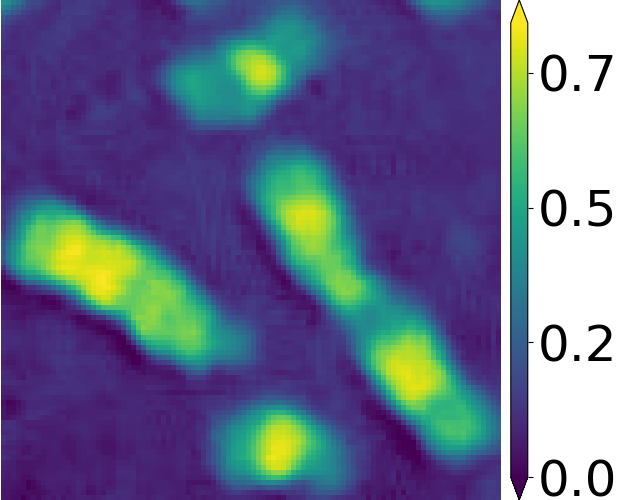}
        \subcaption{Phantom}
        \label{fig:chrom-noise:ph}
    \end{subfigure}
    \vskip\baselineskip 
    \centering
    \begin{subfigure}[b]{0.65\linewidth}
        \includegraphics[width=\linewidth]{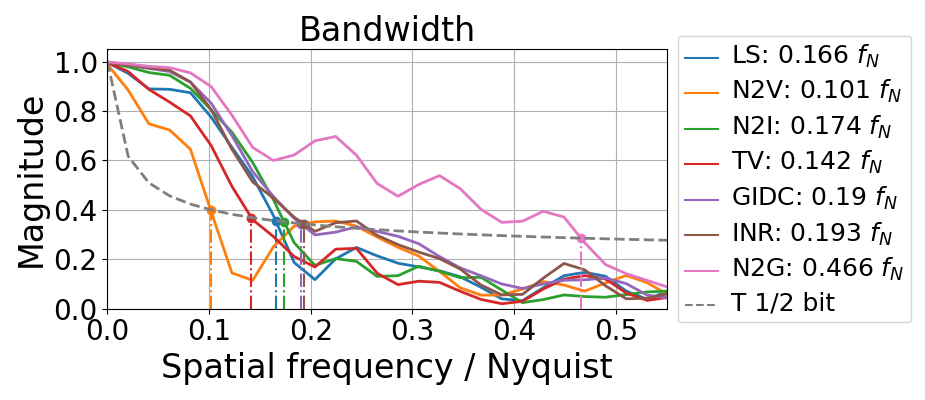}
        \subcaption{FRC}
        \label{fig:chrom-noise:frc}
    \end{subfigure}
    \vskip\baselineskip 
    \begin{subfigure}[b]{0.70\linewidth}
    \begin{center}
        \setlength{\tabcolsep}{3pt} 
        \footnotesize
        \begin{tabular}{|r||c|c|c|c|c|c|c|}
            \hline
             & LS & N2V & N2I & TV & GIDC & INR & N2G \\
            \hline \hline
            MSE $\downarrow$ & 0.150 & 0.040 & 0.042 & 0.047 & 0.013 & 0.012 & \textbf{0.006} \\
            \hline
            PSNR $\uparrow$ & 6.75 & 12.53 & 12.27 & 11.81 & 17.29 & 17.79 & \textbf{20.58} \\
            \hline
            SSIM $\uparrow$ & 0.034 & 0.122 & 0.142 & 0.186 & 0.337 & 0.380 & \textbf{0.693} \\
            \hline
            Resolution $\downarrow$ & 12.1 & 19.7 & 11.5 & 14.1 & 10.5 & 10.3 & \textbf{4.3} \\
            \hline
        \end{tabular}
        \subcaption{Performance comparison.}
        \label{fig:chrom-noise:table}
    \end{center}
    \end{subfigure}

    \caption{GI reconstructions of the chromosomes phantom, with 3$\times$ compression ratio and very high Poisson noise: Mean noise fluctuations $\sim 164.44\%$ of the mean clean bucket fluctuations. From~(\textbf{\subref{fig:chrom-noise:ls}}) to~(\textbf{\subref{fig:chrom-noise:n2g}}) we show the reconstructions of~(\subref{fig:chrom-noise:ph}) with LS, TV, GIDC, INR and N2G respectively. In~(\textbf{\subref{fig:chrom-noise:frc}}) and in~(\textbf{\subref{fig:chrom-noise:table}}) we present the Fourier ring correlation against~(\textbf{\subref{fig:chrom-noise:ph}}) and various performance metrics of each reconstruction, respectively.}
    \label{fig:chrom-noise}
    \vspace{-0.2cm}
\end{figure}
\\
We present the results of (b) in~\cref{fig:chrom-noise}. Here, we produced 3333 realizations, resulting in a 3$\times$ compression ratio. The applied Poisson noise of 2 maximum emitted photons per pixel per realization (i.e., constant $C = 2$) resulted in fluctuations equal to $164.44\%$ of the mean clean bucket fluctuations. All the reconstructed images are noticeably affected by noise with a larger bandwidth than in the previous examples. N2G can suppress high-frequency noise, but cannot completely correct lower-frequency noise. Despite that, by looking at the FRC in~\cref{fig:chrom-noise:frc}, we notice that it preserves much more bandwidth than the other algorithms and presents a much better resolution.
This again results in the N2G reconstruction having the highest PSNR and SSIM values, and the lowest MSE among all reconstructions. These results are summarized in~\cref{fig:chrom-noise:table}.
\\
As a final remark, we observe that in both the presented cases, N2G tends to provide even smoother backgrounds than the phantom itself. The overly smoothed features are usually high-frequency and low-signal features, which are much more overwhelmed by noise than the high-signal parts of the image. As we discussed in \cref{sec:method}, the suppression of high-frequency noise can result in oversmoothing in these cases.

\subsection{Demonstration with experimental data}
Finally, we compare the reconstruction performance of N2G on real GI data acquired at the beamline ID19 of the ESRF --- The European Synchrotron~\cite{Manni2023a}. The considered dataset consists of $42 \times 87$ pixels images, with $24 \mu$m pixel size, and 896 realizations, where the buckets are the emitted XRF signals from a sample composed of a glass capillary and three wires. Two of the three wires are made of Cu, and they will be the focus of this reconstruction. We used the signal from their $K_\alpha$ emission line. At each realization, we exposed the sample to a 26 keV incoming X-ray beam for 0.1 seconds and acquired each realization 32 times. The accumulated 3.2-second exposures serve as high-SNR buckets. Reducing the number of accumulated exposures artificially reduces the collected number of photons (photon flux), thus simulating lower dose depositions. The XRF signals are collected with a single-pixel hyper-spectral detector, whose spectral output is discretized into 4096 bins of 150 eV energy steps. Each XRF peak has a Lorentzian shape around the mean energy $E_{e,l}$. This means that we observe photon counts in adjacent bins around the expected mean signal energy, which decrease $\sim \gamma_{e,l} / ((E_{e,l} - E)^2 + \gamma_{e,l}^2)$ for bins at energy $E$ further away from it, where $\gamma_{e,l}$ is a constant that depends on the emission line. These neighboring bins are either summed, or the area under their curve is fitted to increase the signal's SNR. Therefore, we can simulate further dose reductions by summing fewer and lower-count bins.
This is especially important if we consider that at the moment there exists only one dataset of nano-scale XRF-GI~\cite{Li2023}. Our dose reduction simulation technique allows us to emulate cases of much lower fluxes such as nano-scale acquisitions.
\begin{figure}[ht]
    \centering
    \begin{subfigure}[]{0.71\linewidth}
        \includegraphics[width=\linewidth]{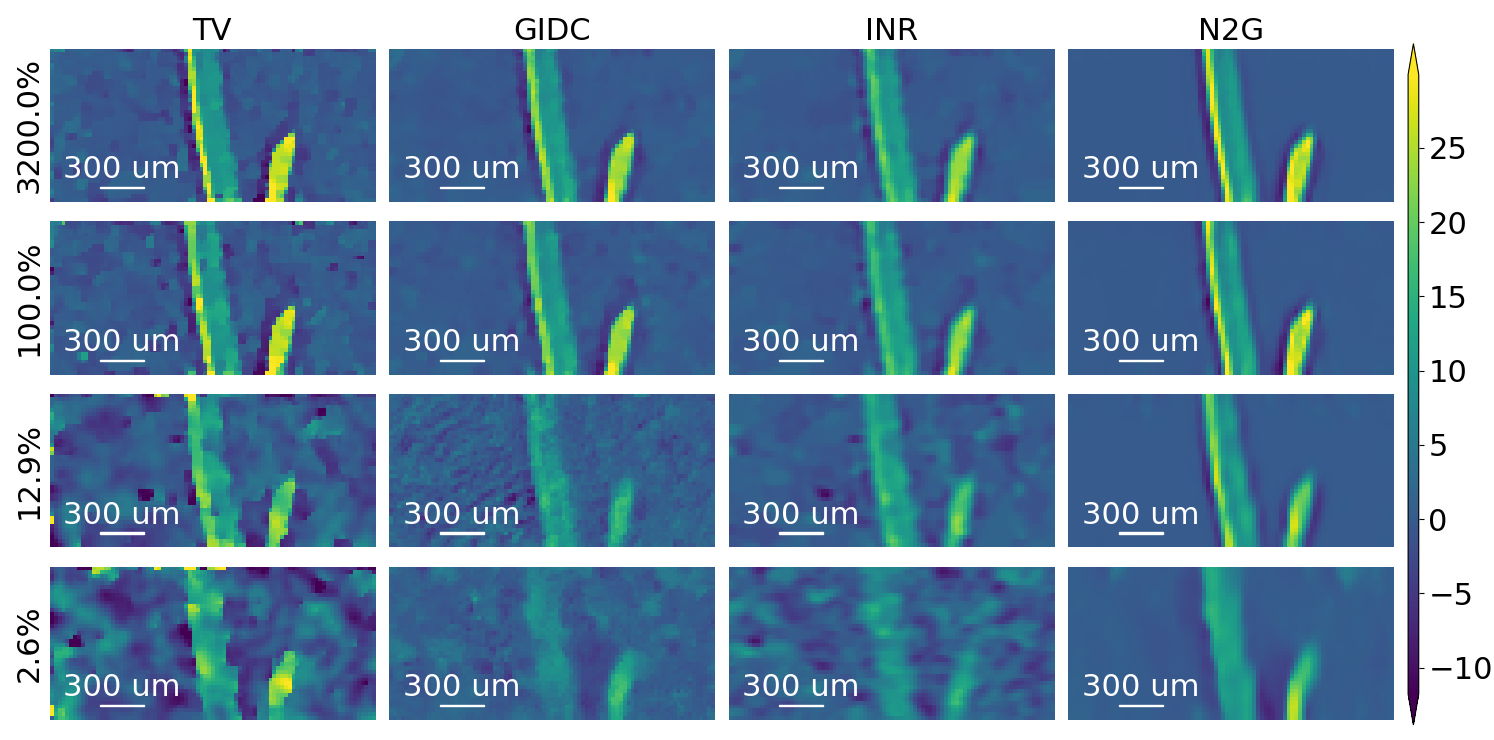}
        \subcaption{XRF-GI reconstructions of two Cu wires.}
        \label{fig:xrf-gi:recs}
    \end{subfigure}
    \hfill
    \begin{minipage}{0.27\linewidth}
        \begin{subfigure}[]{\linewidth}
            \includegraphics[width=\linewidth]{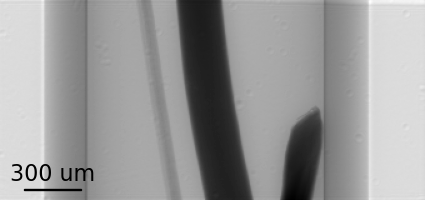}
            \subcaption{Transmission image.}
            \label{fig:xrf-gi:trans}
        \end{subfigure}
        \vskip\baselineskip 
        \begin{subfigure}[]{\linewidth}
            \includegraphics[width=\linewidth]{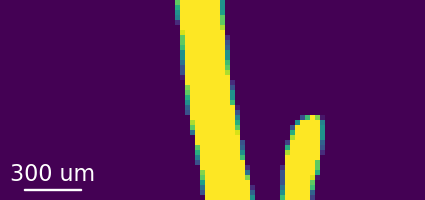}
            \subcaption{Segmented Cu wires.}
            \label{fig:xrf-gi:seg}
        \end{subfigure}
    \end{minipage}
    \vskip\baselineskip 
    \begin{subfigure}[b]{0.52\linewidth}
    \begin{center}
        \setlength{\tabcolsep}{2pt} 
        \footnotesize
        \begin{tabular}{|c||c|c|c|c|}
            \hline 
            Avg. photons & 32000\% & 100\% & 12.9\% & 2.6\% \\
            \hline \hline
            Bucket & 2.07e+05 & 6.47e+03 & 8.38e+02 & 1.69e+02 \\
            \hline
            Bucket \& pixel & 3.48 & 0.109 & 0.0141 & 0.00283 \\
            \hline 
        \end{tabular}
        \subcaption{Photon counts of each acquisition.}
        \label{fig:xrf-gi:photons}
    \end{center}
    \end{subfigure}
    \hfill
    \begin{subfigure}[b]{0.46\linewidth}
    \begin{center}
        \setlength{\tabcolsep}{2pt} 
        \footnotesize
        \begin{tabular}{|c||c|c|c|c|c|c|}
            \hline
            & \multicolumn{3}{c|}{\textbf{PSNR}} & \multicolumn{3}{c|}{\textbf{SSIM}} \\
            & 100\% & 12.9\% & 2.6\% & 100\% & 12.9\% & 2.6\% \\
            \hline \hline
            TV & 30.24 & 24.34 & 21.19 & 0.680 & 0.319 & 0.192 \\
            \hline 
            GIDC & 32.76 & 23.23 & 21.51 & 0.721 & 0.315 & 0.217 \\
            \hline 
            INR & 33.14 & 24.73 & 19.43 & 0.771 & 0.389 & 0.182 \\
            \hline 
            N2G & \textbf{35.33} & \textbf{27.75} & \textbf{23.76} & \textbf{0.933} & \textbf{0.796} & \textbf{0.501} \\
            \hline 
        \end{tabular}
        \subcaption{PSNR and SSIM of the GI reconstructions.}
        \label{fig:xrf-gi:metrics}
    \end{center}
    \end{subfigure}

    \caption{Real data XRF-GI reconstructions: (\textbf{\subref{fig:xrf-gi:recs}})~reconstructions at different photon counts (and corresponding noise levels); (\textbf{\subref{fig:xrf-gi:trans}})~and (\textbf{\subref{fig:xrf-gi:seg}})~an x ray transmission image of the sample at 26 keV and the segmentation of the Cu wires, respectively; (\textbf{\subref{fig:xrf-gi:photons}})~the photoncounts of each acquisition; and (\textbf{\subref{fig:xrf-gi:metrics}})~PSNR and SSIM of each reconstruction against their high-SNR versions in the top row of~(\textbf{\subref{fig:xrf-gi:recs}}).}
    \label{fig:xrf-gi}
    \vspace{-0.2cm}
\end{figure}
\\
We present the results of the XRF-GI reconstructions in~\cref{fig:xrf-gi}. \Cref{fig:xrf-gi:trans} presents an X-ray transmission image (radiograph) of the sample, while~\cref{fig:xrf-gi:seg} indicates the regions that give rise to the XRF signal of interest.
As mentioned above, we consider the 0.1-second exposures as the ``100\% dose'' exposures, while the 3.2-second accumulated images serve as ground truth (high-quality reference images).
The other two images in \cref{fig:xrf-gi:recs} were produced using the five highest-counts and the single highest-counts XRF bins, which provide an average of 12.9\% and 2.6\% of the reference dose per realization, respectively.
These two reduced-flux datasets could thus correspond to 12.9 ms and 2.6 ms exposures per realization, respectively.
The corresponding average photons per bucket and per bucket \& pixel can be found in \cref{fig:xrf-gi:photons}.
The table from \cref{fig:xrf-gi:metrics} presents the estimated image quality degradation with increasing noise levels, according to PSNR and SSIM. Each result is compared against its own method-related ``reference'' reconstruction because we lack a PB scan of the object, which would serve as shared ground truth. Despite that, N2G provides the highest noise reduction according to PSNR and SSIM across all the reconstruction methods considered.

\section{Conclusions \& Outlook}
In this article, we present a self-supervised deep-learning GI reconstruction method specifically targeted at dealing with random noise in the acquired data. If provided with enough high-quality reference data on the studied samples, existing supervised approaches might exhibit high noise suppression performance, with possibly higher PSNR and/or resolution than our approach. However, gathering high-quality reference data can be expensive and requires extensive and organized efforts, thus sometimes proving to be a difficult or even impossible task.
This particularly applies to cutting-edge and niche applications in micro- and nano-imaging, dealing with rare (or even unique) samples with high radiation-dose sensitivity (e.g., biological specimens, battery cells) in in-vivo/in-operando conditions.
In those cases, high-quality reference data are usually simply unavailable. Our approach solves this problem with self-supervision, resulting in higher reconstruction performance against the state-of-the-art unsupervised approaches across multiple metrics like MSE, PSNR, SSIM, and image resolution.
From both the synthetic test cases and the real data reconstructions, we observe that our method preserves higher reconstruction quality than existing unsupervised methods as the noise increases, i.e., at faster acquisition speeds and/or lower deposited dose, across a suite of case studies. This translates into the following relevant aspects: For the same image quality, we could be looking at a reduction in the acquisition time and/or delivered dose by a factor of $\sim$ 1.5 -- 2, compared to the best existing unsupervised methods, when working with the lower deposited doses; and the range of deposited doses where GI offers an advantage over PB is also expanded.
\\
The above-mentioned aspects also mean that our method could enable operating GI scans at $\sim$ 1.5 -- 2 shorter dwell time, resulting in faster GI acquisitions. This aspect is crucial for real case scenarios in two alternative ways: Either increasing acquisition stability or the ability to acquire more realizations for the same dose. In the first case, when paired with the reduced number of GI realizations compared to PB acquisitions, we further reduce the total acquisition times of GI acquisitions, strongly mitigating the impact of sample drifts in our GI scans. Moreover, compared to PB scans, where the sample moves across the field of view, in a GI acquisition, the sample is fixed~\cite{Manni2023}. This further consolidates GI as a very attractive technique for reducing sample drifts and positioning errors in nano-scale imaging. Alternatively, the reduced exposure time for each GI realization would also enable more realizations to be acquired within a fixed acquisition time budget. This, in turn, is the only way to reduce the missing-realization noise, which is the ultimate limit in GI reconstruction quality, as seen earlier in the text.
\\
N2G focuses on the reduction of noise from acquired measurements. It delegates dealing with the missing realization artifacts to the generalization power of the underlying model (NN) and the rather simplistic prior knowledge given by the TV term.
Future work could tackle this problem with a combination of a few possible approaches, including using stronger or more complex priors than TV (e.g., multi-level undecimated wavelet minimization) and multi-channel/multi-modal information from correlated signals (e.g., the transmission signal in~\cite{Li2023}), as demonstrated in computed tomography~\cite{Rigie2015}.

\begin{backmatter}
\bmsection{Funding} Placeholder text for the funding, as required by the submission system.


\bmsection{Disclosures} The authors declare no conflicts of interest.

\bmsection{Data Availability Statement} Data underlying the experimental results are available at~\cite{Manni2023a}.

\bmsection{Supplemental document} See Supplement 1 for supporting content.

\end{backmatter}


\bibliography{references}

\end{document}